\documentclass[letterpaper, 10 pt, journal, twoside]{IEEEtran}

\IEEEoverridecommandlockouts
\makeatletter
\let\NAT@parse\undefined
\makeatother

\usepackage[table]{xcolor}
\usepackage{graphicx}
\usepackage{tikz} 

\usepackage{amsmath, amssymb, bm}
\usepackage{newtxtext, newtxmath} 

\usepackage{booktabs, threeparttable, multirow, array}
\usepackage{dblfloatfix}
\usepackage{subcaption}
\usepackage[font=footnotesize, labelfont=bf]{caption}
\usepackage{adjustbox}

\usepackage{algorithm}
\usepackage{algorithmic}

\usepackage{balance}
\usepackage{soul}
\usepackage{lipsum}
\usepackage{url}
\usepackage[hidelinks]{hyperref}
\usepackage{etoolbox}

\usepackage{rpm_packages/rpm_SIunits}
\usepackage{rpm_packages/rpm_acronyms}
\usepackage{rpm_packages/rpm_math}
\usepackage{rpm_packages/rpm_misc}

\newcommand{\bl}[1]{{\textcolor{black}{#1}}}

\usepackage[numbers]{natbib}
\let\oldthebibliography\thebibliography
\renewcommand{\thebibliography}[1]{
  \oldthebibliography{#1}
  \setlength{\parskip}{0pt}
  \setlength{\itemsep}{-2pt}
}

\title{Cross-Spectral Stereo Inertial Odometry}

\author{
    Seungsang Yun$^{1}$, 
    Hyunsoo Jang$^{1}$, 
    Tai Hyoung Rhee$^{1}$, 
    Hyunho Song$^{1}$, 
    Hyeonjae Gil$^{1}$ 
    and Ayoung Kim$^{1*}$
    \thanks{This work is supported by the Korea Agency for Infrastructure Technology Advancement (KAIA) grant funded by the Ministry of Land, Infrastructure and Transport (Grant RS-2023-00250727) through the Korea Floating Infrastructure Research Center at Seoul National University.}

    \thanks{$^{1}$ S. Yun, H. Jang, T. Rhee, H. Song, H. Gil, and A. Kim are with the Dept. of Mechanical Engineering, Seoul National University (SNU), Seoul, S. Korea. {\tt\footnotesize [seungsang, bronto3082, williamrhee, hun1021405, h.gil, ayoungk]@snu.ac.kr}. Corresponding author: A. Kim}
}

\begin{document}

\maketitle

\begin{abstract}
Standard stereo VIO focuses exclusively on the benefit of metric scale via single-spectrum baselines, often overlooking the risks of \emph{spectral redundancy}. This structural limitation leads to correlated failures, where both sensors simultaneously fail in degraded environments that affect their shared spectrum.
Leveraging a cross-spectral system presents a complementary solution to this issue, yet the significant appearance gap between modalities renders standard matching ineffective. Existing deep learning-based matchers, while effective, introduce inference latencies that violate real-time constraints. To bridge this gap, we present an asynchronous real-time cross-spectral \ac{VTI} system that temporally decouples high-latency deep matching from high-rate state estimation. Our architecture incorporates a spectral-aware weighting scheme that dynamically balances modality reliance based on photometric entropy and thermal noise, ensuring robustness against both abrupt lighting changes and thermal artifacts. Furthermore, we introduce a seamless handling mechanism for thermal \ac{NUC} to maintain tracking continuity. Extensive experiments across diverse scenarios confirm that our system overcomes spectral redundancy, yielding superior accuracy in nominal daylight while ensuring robustness in visually degraded environments.
We will open source our code and data: \bl{\href{https://github.com/seungsang07/cross-spectral-stereo-inertial-odometry} {https://github.com/seungsang07/cross-spectral-stereo-inertial-odometry}}
\end{abstract}
\vspace{-2mm}

\begin{IEEEkeywords} 
Cross-spectral Odometry, SLAM, Sensor fusion
\end{IEEEkeywords}
\vspace{-6mm}

\section{Introduction}
\label{sec:introduction}

\begin{figure}[!t]
    \centering
    \includegraphics[width=0.49\textwidth]{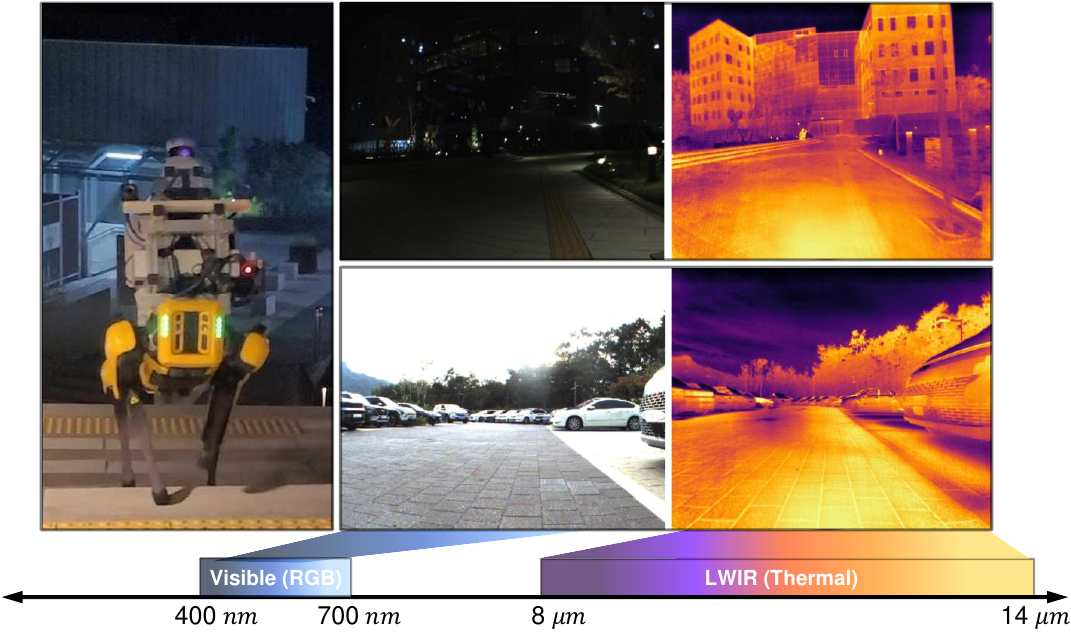}
    \caption{
        Comparison of RGB and \ac{TIR} inputs under challenging illumination. \ac{TIR} preserves structure under low light and glare, enabling robust cross-spectral \ac{VTI} in both degraded and nominal conditions.
    }
    
    \label{fig:main}
    \vspace{-7.0mm}
\end{figure}

Robust state estimation in perceptually degraded environments, such as low-light or abrupt lighting changes, remains a fundamental limitation in practical robotic autonomy. \bl{As shown in Fig.~\ref{fig:main}, RGB and \ac{TIR} images provide complementary cues under both challenging illumination and nominal daytime conditions.} While modern stereo \ac{VIO} has matured into a standard solution, it remains bound by a fundamental structural limitation, which we term \emph{spectral redundancy}, where multiple views capture highly correlated spectral information. Most existing systems rely on homogeneous stereo configurations, where both views operate within the same spectral band. This structural property introduces two critical weaknesses. First, adverse illumination conditions degrade both views in a highly correlated manner, causing homogeneous stereo systems to fail jointly and leading to correlated failure modes where the estimator loses observability. Second, even under nominal conditions, homogeneous stereo suffers from information saturation. Since both views capture identical spectral information, they generate highly correlated photometric gradients. Consequently, instead of sharpening the optimization basin, the second view merely duplicates the existing constraints, offering negligible gain in conditioning. In comparison, cross-spectral stereo configuration fusing different spectral modalities yields sharper and more conditioning-effective constraints, enhancing robustness. 

Beyond improved conditioning, stereo-derived depth remains essential for metric visual-inertial odometry, as it ensures scale observability and helps constrain drift. \bl{\ac{TIR} sensing complements RGB by capturing thermal radiation rather than visible illumination, making RGB-\ac{TIR} odometry attractive under both degraded and nominal lighting conditions.} However, obtaining reliable RGB-\ac{TIR} correspondences is nontrivial, since the substantial appearance gap between modalities violates the photometric consistency or descriptor invariance assumptions underlying classical stereo methods. To address this limitation, recent deep learning-based matchers have been explored to bridge the RGB-TIR appearance gap, but their inference latency ($>$ \unit{500}{ms}) remains prohibitive for real-time estimation. As a result, only a few cross-spectral inertial odometry methods exist \cite{beauvisage2022rmvtio,flemmen2021rovtio}, and most avoid direct RGB-\ac{TIR} matching by decoupling the modalities and relying on indirect scale constraints.

In this work, we reconcile this conflict with a novel tightly-coupled \ac{VTI} odometry framework with asynchronous deep cross-spectral matching. To the best of our knowledge, this is the first work to enable real-time direct RGB-\ac{TIR} stereo matching within a tightly-coupled state estimation pipeline. Our key insight is to \emph{temporally decouple} high-latency deep matching from high-rate state propagation. Instead of stalling the estimator to wait for cross-spectral correspondences, our architecture treats deep matching as an asynchronous auxiliary process that opportunistically reinforces state estimation by injecting metric depth constraints, without interfering with high-rate tracking. To make this practical, we carefully implement a tensor-optimized inference engine that reduces matching latency by $4\times$, ensuring that these cross-spectral metric anchors arrive frequently enough to correct scale drift.

Furthermore, fusing \ac{TIR} imagery introduces modality-specific challenges due to its radiometric characteristics, including sensor artifacts such as \ac{FPN} and intermittent \ac{NUC}-induced data freezes that can destabilize optimization if fused naively. We address these effects using a novel \ac{SRW} scheme to down-weight unreliable thermal measurements based on information-theoretic and wavelet-based noise analysis, together with a dedicated \ac{NUC} handling mechanism to preserve estimator consistency during freezes.

%
The primary contributions of our paper are:

\begin{itemize}
    \item The first geometrically tightly-coupled cross-spectral stereo \ac{VTI} framework that asynchronously integrates tensor-level accelerated deep cross-spectral matching as opportunistic constraints, enabling robust metric scale recovery without stalling real-time state estimation.
    \item A novel \emph{Spectral Reliability Weight (SRW)} metric that robustly fuses cross-spectral sensors by dynamically modulating residuals based on local information density and thermal noise.
    \item Extensive validation on diverse robotic platforms, demonstrating superior robustness against spectral redundancy compared to state-of-the-art single-modality baselines. Code and dataset will be released for the community.
\end{itemize}

\vspace{-4mm}

\section{Related Works}
\label{sec:related_works}

\subsection{Homogeneous Stereo Configuration}

\subsubsection{Visible Spectrum Stereo}
Homogeneous stereo visual-inertial odometry has become a standard foundation for metric motion estimation by tightly coupling stereo vision and inertial sensing within a unified estimator. Representative systems include VINS-Fusion \cite{qin2018online_vinsfusion}, which extends filter-based visual-inertial estimation to multi-sensor configurations, ORB-SLAM3 \cite{campos2021orbslam3}, which provides a complete SLAM framework with robust initialization, and OKVIS2 \cite{leutenegger2022okvis2}, which introduces scalable keyframe-based optimization. DM-VIO \cite{von2022dmvio} extends the direct sparse formulation of DSO \cite{engel2018dso} to the visual-inertial domain, further improving consistency through delayed marginalization. However, relying exclusively on pairs of homogeneous sensors creates a fundamental vulnerability: \textit{spectral redundancy}. Under adverse conditions, both sensors degrade simultaneously, leaving the estimator prone to correlated failure modes regardless of the algorithmic sophistication.

\subsubsection{Thermal Spectrum Stereo}
Homogeneous stereo formulations have also been explored for the thermal spectrum to address low-light conditions. However, thermal imagery suffers from low texture, weak gradients, and a high susceptibility to sensor noise, which undermine the repeatability of handcrafted features and direct photometric assumptions. While recent approaches leverage learned representations to stabilize thermal tracking \cite{saputra2020deeptio,zhao2020tptio}, thermal-only stereo remains substantially less precise than RGB in nominal conditions and fails to exploit the complementary nature of multi-modal sensing.


\subsection{Multi-Spectral Configuration}

\subsubsection{Cross-Spectral Correspondence}
Cross-spectral matching is inherently challenging because visible-thermal pairs violate photometric consistency and exhibit radiometric instability. Early works relied on modality-invariant measures such as Mutual Information (MI) or handcrafted descriptors paired with global regularization \cite{krotosky2009registering}. While these approaches partially mitigate appearance discrepancies, they remain brittle in low-texture scenes and susceptible to thermal noise. Recent deep learning-based methods have substantially narrowed the modality gap. Approaches like MultiPoint \cite{achermann2021multipoint} learn shared keypoints, while large-scale transformer matchers such as XoFTR \cite{tuzcuouglu2024xoftr} and MINIMA \cite{ren2025minima} achieve impressive robustness. 
However, a critical gap remains in their deployment: these deep matchers introduce prohibitive inference latencies, making them computationally incompatible with the real-time constraints. Our work bridges this gap by adopting an asynchronous architecture with tensor-level acceleration, enabling the practical use of dense cross-modal matching without stalling high-rate tracking.

\subsubsection{Multi-Spectral Stereo Odometry}
Early multi-spectral stereo odometry sought to recover metric motion by pairing visible and thermal cameras but quickly exposed the brittleness of cross-spectral correspondence. The seminal Multispectral Stereo Odometry \cite{mouats2014multispectral_stereo_odometry} demonstrated feasibility under controlled conditions, yet relied on handcrafted descriptors sensitive to radiometric variability. Consequently, modern systems often sidestep explicit cross-spectral stereo matching in favor of looser coupling. Dai \textit{et al.} \cite{dai2019multispectral_vo} rely on direct multi-view alignment without stereo matching, while ROVTIO \cite{flemmen2021rovtio} leverages both visible/thermal spectrum but explicitly avoids cross-spectral stereo correspondences, treating the system as effectively monocular per modality. A similar strategy is adopted in \cite{beauvisage2022rmvtio}, where the two sensors are processed separately without explicit cross-spectral matching and later integrated using mutual information.
Other multimodal pipelines emphasize robustness via modality switching or heuristic fusion \cite{beauvisage2020multimodal_tracking_vo}.

Despite meaningful progress, these systems fail to fully exploit the geometric benefits of stereo, such as instantaneous scale recovery. In contrast, our approach represents the first tightly-coupled \ac{VTI} framework to integrate dense, deep cross-spectral matching into online optimization, completing stable cross-spectral metric depth.

\vspace{-6mm}

\section{Direct Visual-Thermal-Inertial Odometry}
\label{sec:methods}
\subsection{System Overview}
\label{sec:system_overview}
\begin{figure}[!t]
    \vspace{1.0mm}
    \centering
    \includegraphics[width=0.495\textwidth]{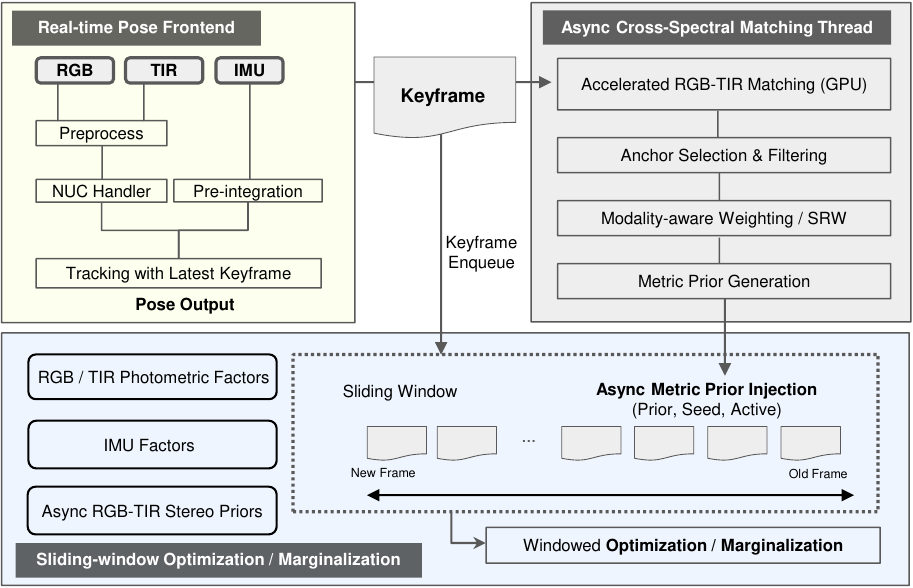}
    \caption{
        \textbf{System overview} of the proposed real-time \ac{VTI} odometry. RGB, \ac{TIR}, and IMU measurements are processed by a \bl{non-blocking frontend for real-time pose output}. \bl{Selected keyframes trigger asynchronous RGB-\ac{TIR} matching, and the resulting SRW-weighted metric priors are incorporated into landmark inverse-depth optimization without blocking the frontend.} During NUC dropouts, \bl{\ac{TIR} factors and RGB-\ac{TIR} priors are skipped,} enabling RGB-IMU fallback. \bl{Runtime statistics are reported in Table~\ref{tab:runtime_stats}.}
    }
    \label{fig:flow_chart}
    \vspace{-5.0mm}
\end{figure}

As illustrated in \figref{fig:flow_chart}, we present a tightly-coupled cross-spectral stereo-\ac{VTI} system that reconciles high-latency deep matching with real-time state propagation. Deep matching runs asynchronously to avoid stalling the high-rate tracking loop, yet its RGB-TIR correspondences are weighted by SRW and \bl{incorporated into the sliding-window estimation pipeline through metric inverse-depth priors, preserving tight coupling with the state.} We further adopt a dual-stream \ac{TIR} strategy: Fieldscale~\cite{gil2024fieldscale} enhances saliency for matching, while 14-bit raw measurements preserve photometric consistency for tracking.


\vspace{-2mm}

\subsection{Spectral Reliability Weight}
\label{sec:ewg_filter}
In the front-end of a cross-spectral odometry pipeline, treating measurements from different spectral bands with uniform reliability is suboptimal due to their disparate radiometric characteristics. RGB imagery is mainly degraded by external illumination, whereas \ac{TIR} imagery is affected by emissivity, heat diffusion, and sensor-specific artifacts such as \ac{FPN}. As a result, image gradients are not always equally indicative of stable geometric structure, and naive cross-spectral fusion can lead to biased or unstable alignment.

To address this issue, we first modulate each visual and thermal residual using a lightweight reliability weight $\xi$ based on a point-wise extension of \ac{EWG}~\cite{kim2020proactive}. Given a grayscale image $I^{\mathrm{gray}}\!\in[0,1]$, we define a per-pixel reliability weight $\xi^\text{img}$ for point $x_i$ as:
\vspace{-2mm}
\begin{equation}
    \begin{aligned}
    \xi_i^{\mathrm{img}}(x_i)
    =
    \max\!\Big\{&
    \mathbb{I}\!\left[\Gamma(x_i)\ge 0.01\,\mathbb{E}[\Gamma]\right]\,
    \Gamma(x_i) \\
    &-
    \mathbb{I}\!\left[H_{\mathrm{bits}}(x_i)\le\tau_H\right]\,
    \mathbb{E}[\Gamma],
    \;0
    \Big\}
    \label{eq:ewg}
    \end{aligned}
\vspace{-2mm}
\end{equation}

\noindent where $x_i$ denotes the image location of the selected point $i$, $\Gamma(x_i)$ denotes the Sobel gradient magnitude at $x_i$, $H_{\mathrm{bits}}(x_i)$ is the local Shannon entropy within a small patch centered at $x_i$, \bl{and $\tau_H$ is the entropy threshold for identifying low-information regions.}

\bl{
However, the point-wise reliability $\xi^{\mathrm{img}}(x_i)$ focuses on local informativeness from gradient magnitude and entropy. This is insufficient for \ac{TIR} images: when the scene-level thermal contrast is weak, local responses become more susceptible to sensor-induced perturbations such as \ac{FPN} or thermal noise. We therefore complement the point-wise reliability with a frame-level Haar-based reliability term, HNR, which measures the amount of fine-scale thermal response relative to the coarse, low-frequency background variation. A high HNR indicates that the frames contain rich scene-dependent detail, while a low HNR attenuates residuals from weakly structured thermal frames.
}

\begin{figure}[!t]
\vspace{1.0mm}
    \centering
    \includegraphics[width=0.475\textwidth]{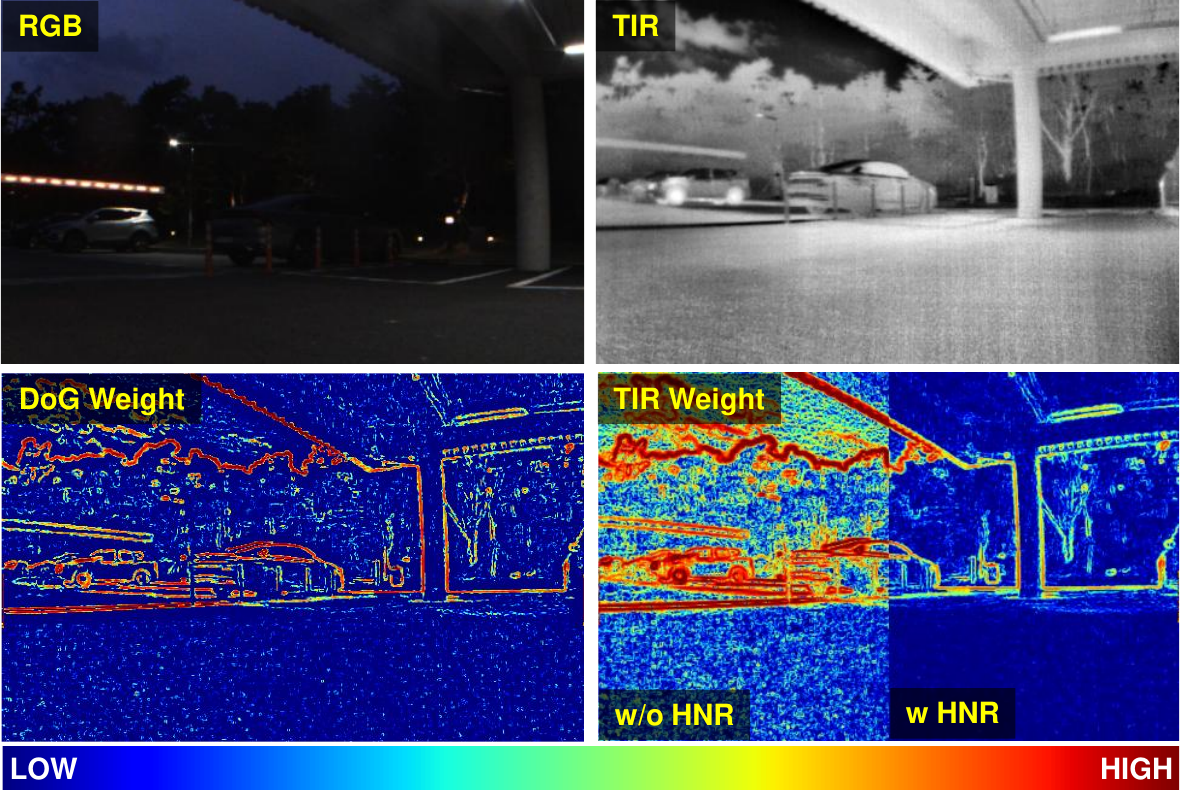}
    \caption{
        \textbf{SRW map of RGB-\ac{TIR} measurements.} Higher values denote greater photometric reliability. \bl{The proposed \ac{SRW} combines point-wise EWG with frame-level HNR to suppress noise-dominated \ac{TIR} responses while preserving reliable structures. For comparison, the \ac{DoG} response is also visualized; it captures structural edges but also responds to high-frequency thermal noise and FPN-like artifacts.}
    }
    \label{fig:adaptive_weight}
    \vspace{-6.0mm}
\end{figure}

\bl{
Given a $L$-level 2D Haar discrete wavelet transform, let $c_A^{(L)}$ denote the coarsest approximation coefficients and $\{c_H^{(1)},c_V^{(1)},c_D^{(1)}\}$ the fine-level detail coefficients. We define:
}
\begin{equation}
    \bl{
    E_{\mathrm{low}}=\mathbb{E}\!\left[(c_A^{(L)})^4\right], \quad
    E_{\mathrm{high}}=\frac{1}{3}\sum_{\alpha\in\{H,V,D\}}
    \mathbb{E}\!\left[(c_\alpha^{(1)})^2\right].
    }
    \label{eq:haar_energy}
\end{equation}

\noindent
\bl{
and compute the HNR:
}

\begin{equation}
    \bl{
    HNR
    =
    \mathrm{clip}\!\left(
    \frac{E_{\mathrm{high}}}{E_{\mathrm{low}}+\epsilon}\,2^{B}\kappa,\;0,\;1
    \right)
    }
    \label{eq:hnr}
\end{equation}

\noindent
\bl{
where $B$ is the sensor bit depth, $\kappa$ is a fixed scaling constant, $\epsilon$ ensures numerical stability, and $\mathrm{clip}(\cdot,0,1)$ constrains HNR to $[0,1]$. Here, $E_{\mathrm{low}}$ represents low-frequency thermal background variation, while $E_{\mathrm{high}}$ represents Haar detail responses from local thermal boundaries, edges, or fine structures. Their normalized ratio serves as a practical indicator of useful thermal detail relative to the coarse background. A smaller ratio indicates that the frame contains limited structural signal and is therefore more sensitive to radiometric perturbations. \figref{fig:adaptive_weight} shows that HNR attenuates the contribution of weakly structured thermal measurements, thereby reducing the influence of noise-sensitive residuals in the optimization.
}

\bl{
Thus, $\xi^{\mathrm{img}}(x_i)$ evaluates local point-wise informativeness, while HNR provides a frame-level reliability scale for thermal measurements. We introduce \ac{SRW} as their product to stabilize the photometric objective by reducing the contribution of thermal residuals whose structural signal is weak relative to sensor-induced perturbations, while preserving informative thermal structures when available, as demonstrated later in \figref{fig:residual_plot}.
}
\vspace{-2mm}
\begin{equation}
    \bl{
    SRW^\text{img}(x_i) = HNR^{img}\, \xi^{\mathrm{img}}(x_i)
    }
    \label{eq:srw}
\end{equation}


\vspace{-7mm}
\subsection{Cross-spectral Alignment and Correspondence}
\label{sec:rgb_tir_matching}

\subsubsection{Cross-spectral Camera Calibration}
\label{sec:rgb_tir_calib}
We employ a circular pattern-based calibration method~\cite{song2024unbiased} implemented on a custom PCB that actively heats predefined regions to generate thermal contrast. This design facilitates robust feature detection across RGB and \ac{TIR} modalities, enabling accurate cross-spectral alignment. The calibration accuracy is validated through epipolar correspondence, as illustrated in \figref{fig:calib_result}.

\subsubsection{Accelerated Deep Matching via Tensor Optimization}
To recover metric scale and correct drift, we leverage explicit cross-spectral matching using MINIMA~\cite{ren2025minima}, a cross-spectral variant of RoMa~\cite{Edstedt2024roma}, which is robust to the modality gap. Since its native throughput ($\sim$2~Hz) is insufficient for online deployment, we implement a TorchScript-based static execution path. Specifically, we (1) trace the model at a fixed $640 \times 512$ resolution to remove Python overhead and freeze control flow, (2) enable JIT operator fusion to reduce CUDA kernel launches and memory traffic, and (3) reuse pre-allocated buffers for intermediate feature maps to mitigate allocator overhead. On an NVIDIA RTX~4080 GPU, these optimizations yield a $4\times$ speedup. Coupled with asynchronous metric depth injection (\secref{sec:async_depth_injection}), the resulting throughput provides sufficiently frequent metric constraints to preserve global scale consistency. \bl{As reported in \tabref{tab:runtime_stats}, this asynchronous module does not block the pose output.}

\vspace{-4mm}

\subsection{Direct Visual-Thermal-Inertial Odometry Estimator}
\label{sec:ms_stereo_odometry}
Integrating the proposed \ac{SRW} and accelerated deep matching, we formulate a unified tightly-coupled cross-spectral estimator built upon the DM-VIO~\cite{von2022dmvio} architecture. This framework leverages \ac{SRW} to modulate photometric residuals for cross-spectral robustness, while fusing asynchronous metric constraints to enforce global scale within the online pipeline.

\subsubsection{Notation and Sensor Model}
We use $w$ for the world frame, $b$ for the IMU/body frame, and $c\in\{\mathrm{rgb},\mathrm{tir}\}$ for camera frames.
Let $\mathbf{T}_{ab}\in\mathrm{SE}(3)$ denote the rigid transform from frame $b$ to $a$, with rotation $\mathbf{R}_{ab}\in\mathrm{SO}(3)$ and translation $\mathbf{p}_{ab}\in\mathbb{R}^3$. Each camera is rigidly attached to the IMU with known extrinsics $\mathbf{T}_{bc}$. Given a 3D point $\mathbf{P}^w$, its projection into camera $c$ at time $k$ is

\vspace{-4mm}

\begin{equation}
    \mathbf{u}^{c}_{k}=\pi_c\!\left(\mathbf{T}^{k}_{cw}\mathbf{P}^w\right),
\end{equation}
where $\pi_c(\cdot)$ is the calibrated projection model.

\subsubsection{State and IMU Preintegration}
We adopt a standard keyframe-based VIO state~\cite{forster2015imu}:
\begin{equation}
    \mathbf{x}_k=\{\mathbf{R}_{wb}^k,\mathbf{p}_{wb}^k,\mathbf{v}_{wb}^k,\mathbf{b}_g^k,\mathbf{b}_a^k\}.
\end{equation}
Between consecutive keyframes $k$ and $k{+}1$, we form an IMU preintegration factor from all IMU samples in the interval, yielding residual $\mathbf{r}^{\mathrm{imu}}_{k}$ with covariance $\mathbf{\Sigma}^{\mathrm{imu}}_{k}$, which provides high-rate motion constraints in the sliding-window objective.

\subsubsection{Direct Photometric Formulation}
Let $\mathcal{N}_i$ denote the set of pixels within a small patch centered at pixel $\mathbf{u}_i$ in the host keyframe $h(i)$. \bl{Here, $\rho_i$ is the inverse-depth state of point $i$, initialized from RGB-\ac{TIR} triangulation when available and otherwise refined through direct tracking.} For a pixel $\mathbf{u} \in \mathcal{N}_i$ with inverse depth $\rho_i$, its corresponding projected location $\mathbf{u}'$ in the target keyframe $k$ observed by camera $c$ is given by the warping function:
\vspace{-2mm}

\begin{equation}
    \mathbf{u}' = \pi_c\!\left( \mathbf{T}^{k}_{cw} (\mathbf{T}^{h(i)}_{cw})^{-1} \pi^{-1}_c(\mathbf{u}, \rho_i) \right)
    \label{eq:warping}
    \end{equation}
    where $\pi^{-1}_c(\cdot)$ represents the back-projection function that maps a pixel to a bearing vector. 
    The photometric residual is defined as the intensity difference between the host and the warped target pixels:
    \begin{equation}
    r^{c}_{i,k}(\mathbf{u}) = I_k^c(\mathbf{u}') - I_{h(i)}^c(\mathbf{u})
    \label{eq:photo_residual}
\end{equation}

where $I(\cdot)$ denotes the image intensity. We aggregate the \ac{SRW}-weighted photometric residuals over the patch using a robust loss, yielding the following patch cost:
\vspace{-1mm}
\begin{equation}
    \bl{
    E^{c}_{i,k}
    =
    \sum_{\mathbf{u} \in \mathcal{N}_i}
    \rho_{\mathrm{Huber}}\!\left(
    \left\lVert
    \sqrt{w^{c}_{i,k}(\mathbf{u})}
    \, r^{c}_{i,k}(\mathbf{u})
    \right\rVert^2
    \right)
    }
    \label{eq:weighted_patch_cost}
\end{equation}
\vspace{-2mm}

with the pixel-wise weight defined as:
\vspace{-1mm}
\begin{equation}
    w^{c}_{i,k}(\mathbf{u}) = SRW(\mathbf{u})
    \label{eq:weighted_photo}
\end{equation}

\vspace{-1mm}
\bl{Here, $\rho_{\mathrm{Huber}}(\cdot)$ denotes the Huber loss.} Our SRW-based weighting naturally complements this direct formulation by emphasizing pixels with strong gradients while attenuating high-frequency thermal noise that could otherwise destabilize the multi-spectral residual minimization.

\subsubsection{High-rate Tracking Loop}
At the camera rate, the estimator performs patch tracking via the direct photometric formulation above, IMU preintegration, keyframe selection, and incremental nonlinear optimization on the current sliding window. Notably, this loop is non-blocking; it does not wait for deep matching.

\subsubsection{Sliding-window Objective}
Let $\mathcal{W}$ be the set of keyframes in the current window, 
$\mathcal{L}$ the set of active landmarks, 
\bl{and $\mathcal{S}$ the set of valid asynchronous RGB-\ac{TIR} stereo priors in the current window. 
Let $\mathbf{x}_k$ denote the IMU/body state of keyframe $k$, and let $\boldsymbol{\theta}_i$ denote the landmark parameter, represented by inverse depth in our implementation.} 
We minimize a robustified, weighted least-squares objective of the form:
\begin{equation}
    \begin{aligned}
    \min_{\{\mathbf{x}_k\},\{\boldsymbol{\theta}_i\}}
    \;&
    \sum_{k\in\mathcal{W}}
    \left\lVert \mathbf{r}^{\mathrm{imu}}_k \right\rVert^2_{\mathbf{\Sigma}^{\mathrm{imu}}_k}
    +
    \sum_{i\in\mathcal{L}}\sum_{(c,k)\in\mathcal{O}(i)}
    E^{c}_{i,k}
    +
    \bl{
    \sum_{(i,k)\in\mathcal{S}}
    \left\lVert r^{\mathrm{st}}_{i,k} \right\rVert^2_{\Sigma^{\mathrm{st}}_{i,k}}
    }
    \end{aligned}
    \label{eq:full_objective}
\end{equation}
where $\mathcal{O}(i)$ enumerates the target frames observing the patch anchored at $i$ within the window.
The second term includes both RGB and \ac{TIR} photometric costs with modality-aware residual weighting.
\bl{The third term incorporates stereo-derived inverse-depth priors from valid RGB-\ac{TIR} matches, as detailed in Sec.~\ref{sec:stereo_prior_injection}.}

\vspace{-4mm}
\subsection{Asynchronous Cross-spectral Metric Constraint Integration}
\label{sec:async_depth_injection}

Cross-spectral stereo correspondences are obtained by the deep matcher (\secref{sec:rgb_tir_matching}) and arrive asynchronously. To reconcile latency with real-time tracking, our estimator runs a low-rate matching loop in parallel to the high-rate backbone and injects metric constraints when available, without blocking the frontend.

\newcommand{\FigScale}{0.92}

\begin{figure*}[!t]
    \centering
    \vspace{1.5mm}

    \begin{adjustbox}{scale=\FigScale,center}
    \begin{minipage}{\textwidth}

    \begin{minipage}[t]{0.645\textwidth}
        \vspace{0pt}
        \centering
        \includegraphics[width=\linewidth, trim=0.9mm 0 0 0, clip]{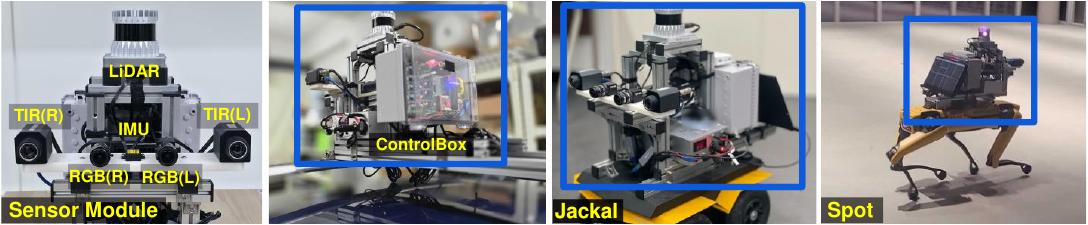}
        \caption{
            \textbf{Hardware setup.} Our custom-built sensor module integrates stereo RGB and \ac{TIR} cameras, an IMU, and a 3D LiDAR. The module uses a custom-built control box that ensures hardware-level synchronization across the four cameras via PWM signals. To demonstrate the system's adaptability, the same module is deployed across diverse robotic platforms, including a wheeled and a quadrupedal robot.
        }
        \label{fig:sensor_system}
    \end{minipage}
    \hfill
    \begin{minipage}[t]{0.335\textwidth}
        \vspace{0pt}
        \centering
        \includegraphics[width=\linewidth]{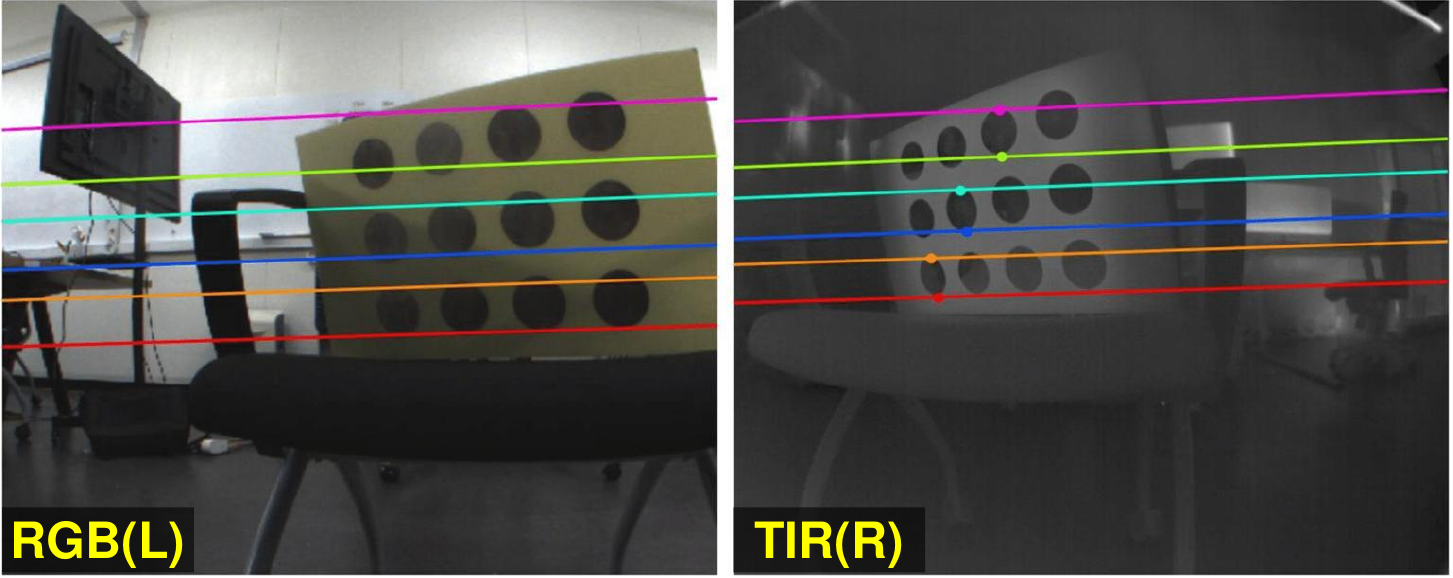} 
        \caption{
            \textbf{Cross-spectral stereo calibration.} Epipolar lines induced by \ac{TIR} points accurately correspond to their RGB counterparts, validating the estimated cross-spectral stereo geometry.
        }
        \label{fig:calib_result}
    \end{minipage}

    \vspace{2mm}
    
    \begin{minipage}[t]{0.60\textwidth}
        \vspace{0pt}
        \centering
        \includegraphics[width=\linewidth]{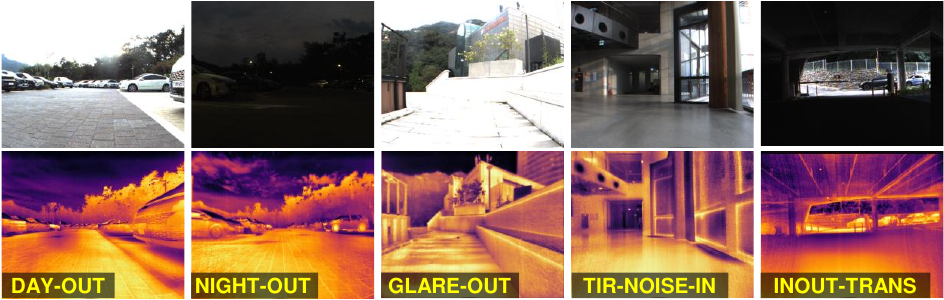}
        \caption{
            \textbf{Sequences from our Cross-Spectral Dataset.} The sequences cover diverse conditions, ranging from nominal lighting to darkness, glare, thermal noise, and illumination changes.
        }
        \label{fig:dataset_samples}
    \end{minipage}
    \hfill
    \begin{minipage}[t]{0.38\textwidth}
        \vspace{0pt} 
        \centering
        \captionof{table}{
            \textbf{Data Sequence Properties.} We evaluated our \ac{VTI} system under various environmental conditions and modality-specific challenges: \textbf{Out} (Outdoor), \textbf{In} (Indoor), \textbf{DRK} (Dark), \textbf{GLR} (Glare), \textbf{FPN} (Fixed Pattern Noise), and \textbf{L.Tex} (Low Texture).
        }
        \label{table:tab_seq_description}
        
        \renewcommand{\arraystretch}{0.9}
        \setlength{\tabcolsep}{3pt}
        
        \resizebox{\linewidth}{!}{
            \begin{tabular}{l|c| c c | c c c c}
            \toprule
              &   & \multicolumn{6}{c}{\textbf{Challenge Factors}} \\
            \cmidrule(l){3-8}
            \textbf{Sequence} & \textbf{Len} & \textbf{Out} & \textbf{In} & \textbf{DRK} & \textbf{GLR} & \textbf{FPN} & \textbf{L.Tex} \\
            \midrule
            \texttt{DAY-OUT}       & 104 m & O &   &   &   &   &   \\
            \texttt{NIGHT-OUT}     & 58 m  & O & O & O &   & O &   \\
            \texttt{GLARE-OUT}     & 67 m  & O &   &   & O &   &   \\
            \texttt{TIR-NOISE-IN}  & 64 m  &   & O &   & O & O & O \\
            \texttt{INOUT-TRANS}   & 227 m & O & O & O & O &   &   \\
            \bottomrule
            \end{tabular}
        }
    \end{minipage}

    \end{minipage}
    \end{adjustbox}

    \vspace{-6mm}
\end{figure*}

\subsubsection{Asynchronous Matching and Constraint Generation}
\label{sec:async_stereo_constraints}
When a new keyframe $k$ is created, we enqueue synchronized RGB/TIR images to the GPU-based matcher. When the matcher returns, it provides correspondences $\{(\mathbf{u}^{\mathrm{rgb}}_{i,k},\mathbf{u}^{\mathrm{tir}}_{i,k})\}$ that satisfy epipolar consistency under the calibrated extrinsics. We triangulate each match to obtain a metric inverse-depth estimate $\hat{\rho}_{i,k}$. These estimates are injected into the active window as additional constraints that sharpen scale observability and improve conditioning.

\subsubsection{Stereo Depth Injection as a Metric Prior}
\label{sec:stereo_prior_injection}
For a matched landmark, we treat the triangulated cross-spectral estimate $\hat{\rho}^{\mathrm{st}}_i$ as a metric measurement of its inverse depth.
\bl{This prior is incorporated into the sliding-window objective as the stereo prior term in Eq.~\eqref{eq:full_objective}.}
We define a lightweight stereo prior:
\vspace{-2mm}
\begin{equation}
     \bl{r^{\mathrm{st}}_{i} = \rho_i - \hat{\rho}^{\mathrm{st}}_i}
\label{eq:stereo_prior}
\end{equation}

\bl{This term guides the corresponding landmark inverse depth toward the metric estimate obtained from RGB-\ac{TIR} triangulation.}
This improves scale conditioning with little computation and allows delayed priors to be incorporated without rewinding the estimator.

\subsubsection{Metric Anchor Selection and Robustification (Prior/Seed/Active)}
\label{sec:point_budgeting}
Cross-spectral matching is prone to outliers, and even a few mismatches can bias metric priors and destabilize scale. We therefore separate points by reliability and use them differently. \emph{Prior} points are top-quality correspondences that pass all geometric and consistency filters and serve as metric \emph{anchors}; importantly, only prior points act as scale anchors that maintain the odometry's metric scale. \bl{These filters include epipolar error gating, positive-depth triangulation, valid disparity/depth range checks, and reprojection consistency checks.} \emph{Seed} points are matched but fail at least one anchor filter; they are kept only as candidates for later initialization. Independently, \emph{active} points are selected from high-gradient locations on a coarse grid to ensure wide coverage and robust direct tracking. \bl{To further reduce the influence of remaining outliers, modality-aware weights $w^{c}_{i,k}$ down-weight unreliable measurements before they dominate the photometric optimization, and the Huber robust loss suppresses residual outliers.} This design isolates scale-critical constraints from noisy matches while maintaining tracking robustness.

\subsubsection{Temporal Consistency and Scale-aware Marginalization}
\label{sec:scale_aware_marginalization}
To maintain a bounded computational cost, we employ Schur complement to marginalize old states. Specifically, we implement a scale-aware marginalization policy to address periods of poor matching quality, which can deplete the ratio of active stereo anchors and lead to scale drift. If the number of landmarks with valid stereo priors falls below a predefined threshold, we adaptively delay the marginalization of existing anchors. This ensures that the sliding window always retains a sufficient density of metric constraints to stabilize the scale, even when new reliable matches are temporarily unavailable.

\vspace{-4mm}
\subsection{Thermal NUC Handling and Safe Fallback}
\label{sec:nuc_handling}
\ac{TIR} cameras periodically perform \ac{NUC}, during which the thermal stream can pause for up to $\sim$1.5\,s. Such dropouts are particularly damaging for tightly-coupled odometry and can trigger catastrophic tracking failure, especially under fast rotations where continuity of photometric constraints is critical. We therefore explicitly handle \ac{NUC} to prevent invalid thermal updates from corrupting the estimator. We detect \ac{NUC} events by monitoring cross-spectral synchronization: if only one modality frame arrives, we treat it as a \ac{NUC}-induced dropout and enforce a \ac{NUC}-aware factor policy. Upon detection, we (i) skip constructing \ac{TIR} factors for the affected timestamps, and (ii) reject cross-spectral stereo priors involving the missing/frozen thermal frames. The estimator then safely falls back to RGB-IMU until synchronized RGB-\ac{TIR} updates resume (\figref{fig:flow_chart}).
\vspace{-4mm}

\section{Experiments}
\label{sec:experiments}

\begin{table*}[t]
    \vspace{1.5mm}
    \centering
    \def\TotalWidth{0.98} 

    \begin{minipage}{\TotalWidth\linewidth}
        \centering
        
        \begin{minipage}[c]{0.56\linewidth}
            \centering
            \caption{
                \textbf{Quantitative Comparison on DAY-OUT.} ATE and RTE results are compared against representative stereo \ac{VIO} baselines, including VINS-Fusion \cite{qin2018online_vinsfusion}, ORB-SLAM3 \cite{campos2021orbslam3}, and OKVIS2 \cite{leutenegger2022okvis2}, as well as the cross-spectral baseline ROVTIO \cite{flemmen2021rovtio} \bl{and monocular RGB baseline DM-VIO \cite{von2022dmvio}}. Contrary to the conventional assumption that RGB-stereo is optimal for well-lit environments, our approach demonstrates superior accuracy by overcoming spectral redundancy, \bl{indicating that cross-spectral fusion can provide complementary constraints even in nominal conditions.} \textbf{Bold}: best, \underline{underlined}: second best. \textit{TF} \bl{and \textit{IF} denote Tracking Failure and Initialization Failure, respectively.}
            }
            \label{tab:results_day_out}
    
            \setlength{\tabcolsep}{3.0pt} 
            \renewcommand{\arraystretch}{1.4}
            \setlength{\aboverulesep}{0pt}
            \setlength{\belowrulesep}{0pt}
    
            \resizebox{\linewidth}{!}{%
                \scriptsize
                \begin{tabular}{cc | c ccc ccc c >{\columncolor{black!10}}c}
                \toprule
                \multicolumn{2}{c|}{} 
                & \multicolumn{9}{c}{\texttt{DAY-OUT}} \\
                \cmidrule(lr){3-11}
                \multicolumn{2}{c|}{}
                & \multicolumn{1}{c}{\bl{\textbf{Mono RGB}}}
                & \multicolumn{3}{c}{\textbf{Stereo RGB}} 
                & \multicolumn{3}{c}{\textbf{Stereo TIR}} 
                & \multicolumn{2}{c}{\textbf{Cross-spectral}} \\ 
                \cmidrule(lr){3-3} \cmidrule(lr){4-6} \cmidrule(lr){7-9} \cmidrule(l){10-11}
                \multicolumn{2}{c|}{\textbf{Metric}}
                & \bl{\tiny DM-VIO}
                & \tiny VINS-Fusion & \tiny ORB-SLAM3 & \tiny OKVIS2 
                & \tiny VINS-Fusion & \tiny ORB-SLAM3 & \tiny OKVIS2 
                & \tiny ROVTIO & \tiny \textbf{Ours} \\ 
                \midrule
                \multirow{2}{*}{\textbf{ATE}} 
                  & \textbf{Rot} (deg)
                  & \bl{6.758} & 2.440 & \underline{2.432} & 2.647 & 5.719 & 7.343 & 3.027 & \textit{TF} & \textbf{2.269} \\
                  & \textbf{Trans} (m)
                  & \bl{2.483} & \underline{0.393} & 1.310 & 0.659 & 3.676 & 3.567 & 1.921 & \textit{TF} & \textbf{0.234} \\
                \cmidrule{3-11}
                \multirow{2}{*}{\textbf{RTE}}
                  & \textbf{Rot} (deg)
                  & \bl{0.515} & 0.955 & \underline{0.434} & 1.402 & 0.686 & \textbf{0.422} & 1.281 & \textit{TF} & 0.506 \\
                  & \textbf{Trans} (m)
                  & \bl{0.035} & 0.024 & \underline{0.022} & 0.026 & 0.065 & 0.057 & 0.085 & \textit{TF} & \textbf{0.018} \\
                \bottomrule
                \end{tabular}%
            }
        \end{minipage}
        \hfill
        \begin{minipage}[c]{0.42\linewidth}
            \centering
            \includegraphics[width=\linewidth]{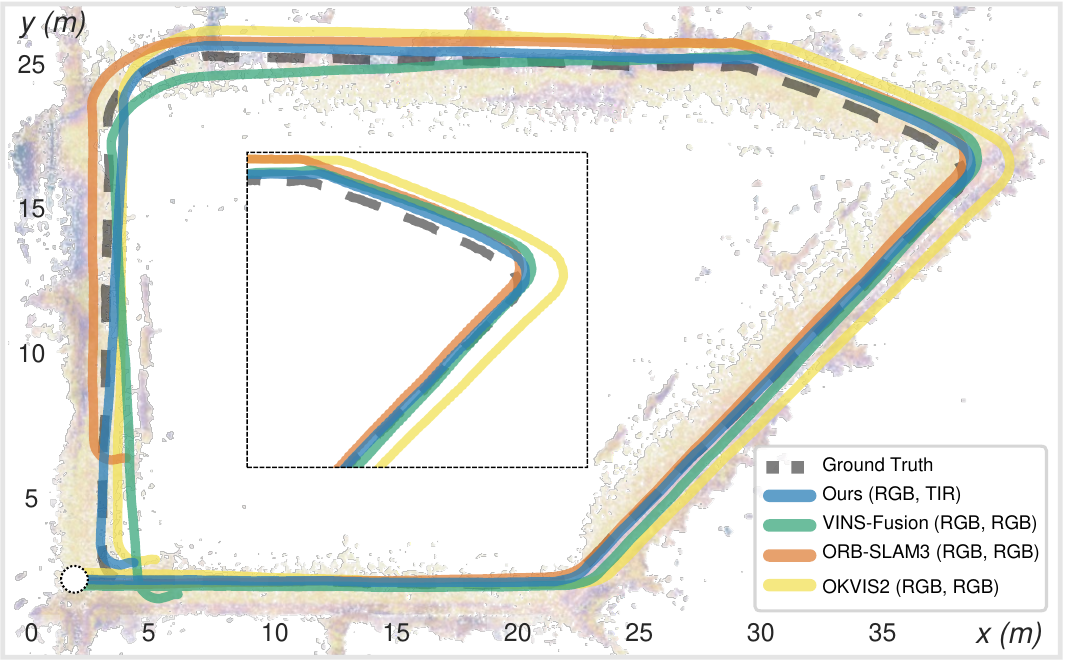}
            
            \captionof{figure}{ 
                \textbf{Qualitative trajectory comparison on DAY-OUT.} Trajectories are overlaid on the reconstructed global \ac{TIR} map \bl{after aligning the initial pose and starting point.}
            }
            \label{fig:trajectory_day_out}
        \end{minipage}
    \end{minipage}
    \vspace{-2mm}    
\end{table*}

\vspace{-2mm}
\begin{table*}[h]
    \centering
    \def\TotalWidth{0.98} 

    \begin{minipage}{\TotalWidth\linewidth}
        \centering
        \caption{ \textbf{Quantitative Comparison on NIGHT-OUT and GLARE-OUT.} ATE and RTE are reported. Best and second-best results are shown in \textbf{bold} and \underline{underlined}, respectively. \textit{TF}/\textit{IF} denote tracking/initialization failure. }
        \label{tab:results_night_glare}

        \setlength{\tabcolsep}{2.5pt} 
        \renewcommand{\arraystretch}{1.4}
        \setlength{\aboverulesep}{0pt}
        \setlength{\belowrulesep}{0pt}

        \resizebox{\linewidth}{!}{%
            \scriptsize
            \begin{tabular}{cc | c ccc ccc c >{\columncolor{black!10}}c | c ccc ccc c >{\columncolor{black!10}}c}
            \toprule
            
            \multicolumn{2}{c|}{} 
            & \multicolumn{9}{c|}{\texttt{NIGHT-OUT}} 
            & \multicolumn{9}{c}{\texttt{GLARE-OUT}} \\
            
            \cmidrule(lr){3-11} \cmidrule(l){12-20}
            
            \multicolumn{2}{c|}{}
            & \multicolumn{1}{c}{\bl{\textbf{Mono RGB}}}
            & \multicolumn{3}{c}{\textbf{Stereo RGB}} 
            & \multicolumn{3}{c}{\textbf{Stereo TIR}} 
            & \multicolumn{2}{c|}{\textbf{Cross-spectral}} 
            & \multicolumn{1}{c}{\bl{\textbf{Mono RGB}}}
            & \multicolumn{3}{c}{\textbf{Stereo RGB}} 
            & \multicolumn{3}{c}{\textbf{Stereo TIR}} 
            & \multicolumn{2}{c}{\textbf{Cross-spectral}} \\ 
            
            \cmidrule(lr){3-3} \cmidrule(lr){4-6} \cmidrule(lr){7-9} \cmidrule(lr){10-11}
            \cmidrule(lr){12-12} \cmidrule(lr){13-15} \cmidrule(lr){16-18} \cmidrule(l){19-20}
            
            \multicolumn{2}{c|}{\textbf{Metric}}
            & \bl{\tiny DM-VIO}
            & \tiny VINS-Fusion & \tiny ORB-SLAM3 & \tiny OKVIS2 
            & \tiny VINS-Fusion & \tiny ORB-SLAM3 & \tiny OKVIS2 
            & \tiny ROVTIO & \tiny \textbf{Ours} 
            & \bl{\tiny DM-VIO}
            & \tiny VINS-Fusion & \tiny ORB-SLAM3 & \tiny OKVIS2 
            & \tiny VINS-Fusion & \tiny ORB-SLAM3 & \tiny OKVIS2 
            & \tiny ROVTIO & \tiny \textbf{Ours} \\ 
            \midrule
            
            \multirow{2}{*}{\textbf{ATE}} 
              & \textbf{Rot} (deg)
              & \bl{\textit{TF}}
              & \textbf{2.159} & 6.648 & \underline{2.522} & 3.451 & 5.439 & 2.926 & 9.718 & 4.315
              & \bl{6.448}
              & 4.297 & 10.144 & 7.028 & \underline{2.270} & 3.392 & 9.869 & 4.638 & \textbf{2.058} \\
             
              & \textbf{Trans} (m)
              & \bl{\textit{TF}}
              & \underline{0.422} & 2.745 & 0.635 & 0.719 & 3.545 & 1.490 & 11.466 & \textbf{0.367}
              & \bl{3.529}
              & \underline{0.699} & 2.363 & 1.658 & 1.037 & 2.386 & 1.893 & 13.348 & \textbf{0.419} \\
            
            \cmidrule{3-20}
            
            \multirow{2}{*}{\textbf{RTE}}
              & \textbf{Rot} (deg)
              & \bl{\textit{TF}}
              & 0.691 & \underline{0.427} & 1.373 & 1.416 & \textbf{0.392} & 1.385 & 0.456 & 0.434
              & \bl{0.838}
              & 1.251 & \underline{0.780} & 1.446 & 1.218 & \textbf{0.736} & 1.355 & 0.803 & 0.810 \\
             
              & \textbf{Trans} (m)
              & \bl{\textit{TF}}
              & \underline{0.032} & 0.043 & 0.101 & 0.050 & 0.079 & 0.111 & 0.244 & \textbf{0.031}
              & \bl{0.058}
              & \underline{0.044} & 0.049 & 0.062 & 0.074 & 0.062 & 0.103 & 0.143 & \textbf{0.026} \\
            \bottomrule
            
            \end{tabular}%
        }
    \end{minipage}
    \vspace{-6mm}
\end{table*}

\subsection{Experimental Setup}
\label{sec:exp_setup}
To evaluate the proposed cross-spectral stereo-inertial system, we utilize a custom modular platform, as shown in \figref{fig:sensor_system}. This platform is equipped with stereo RGB cameras (FLIR Blackfly), stereo \ac{TIR} cameras (FLIR A65), and a MicroStrain 3DM-GX5-25 IMU. All sensors are rigidly mounted and hardware-synchronized via a microcontroller unit that generates PWM trigger signals to ensure precise inter-camera temporal alignment. For reference trajectory generation, a 3D LiDAR is employed exclusively for evaluation purposes. To enable rigorous evaluation across stereo RGB, stereo \ac{TIR}, and RGB-\ac{TIR} cross-spectral stereo configurations, we constructed a comprehensive dataset. \bl{Existing RGB-\ac{TIR} and multi-spectral datasets, including CART~\cite{lee2024cart}, M2P2~\cite{datar2025M2P2}, and MS2~\cite{shin2023CVPR}, mainly target aligned perception, low-light off-road perception, or thermal depth estimation, rather than controlled comparisons among stereo RGB, stereo \ac{TIR}, and RGB-\ac{TIR} stereo under identical geometric conditions.} \bl{We therefore provide} high-precision calibration parameters, validated by the aligned RGB-\ac{TIR} epipolar lines in \figref{fig:calib_result}. As detailed in \tabref{table:tab_seq_description}, the sequences cover diverse adverse conditions. \bl{We report \ac{ATE} and \ac{RTE} using evo~\cite{grupp2017evo} after $\mathrm{SE}(3)$ rigid alignment to the ground truth, without scale correction.} All experiments are performed on a workstation with an Intel Core i9 (2.50\,GHz) CPU, 64\,GB RAM, and an NVIDIA RTX 4080 GPU.

\vspace{-4mm}
\subsection{Baselines}
\label{sec:exp_baselines}
We compare our method against \bl{monocular \ac{VIO} baseline DM-VIO~\cite{von2022dmvio} and} representative stereo \ac{VIO} systems: VINS-Fusion~\cite{qin2018online_vinsfusion}, ORB-SLAM3~\cite{campos2021orbslam3}, and OKVIS2~\cite{leutenegger2022okvis2}. To validate the advantage of cross-spectral fusion over single-modality approaches, we evaluate each stereo baseline in both RGB and \ac{TIR} stereo-inertial configurations. This comparison benchmarks our system against the individual performance of each spectrum. Among existing methods, ROVTIO~\cite{flemmen2021rovtio} is the only publicly available RGB-TIR inertial configuration that allows a direct comparison.

\vspace{-4mm}
\subsection{Quantitative Results}
\label{sec:exp_quant}

Sequence characteristics are summarized in \tabref{table:tab_seq_description}. Detailed quantitative results are presented in \tabref{tab:results_day_out} (nominal condition), \tabref{tab:results_night_glare} (illumination degradation), and \tabref{tab:results_inout_noise} (thermal noise and indoor-outdoor transition).

\paragraph{\texttt{DAY-OUT}}
As shown in \tabref{tab:results_day_out}, this sequence represents a well-lit outdoor scenario where RGB stereo baselines already achieve strong performance due to abundant texture and stable illumination. In such nominal conditions, homogeneous RGB stereo is commonly considered the optimal configuration. Accordingly, it is generally assumed to be sufficient. Nevertheless, our cross-spectral system achieves the lowest translational and rotational \ac{ATE}, outperforming all RGB-only baselines. \bl{ROVTIO failed shortly after the beginning of this sequence. This behavior is likely associated with initialization sensitivity under near-forward motion, since its loosely coupled formulation does not explicitly inject stereo-based metric depth anchors.}

\begin{figure*}[t]
    \centering
    
    \def\TotalWidth{0.98} 

    \begin{minipage}{\TotalWidth\linewidth}
        \centering
        
        \begin{minipage}[t]{0.365\linewidth}
            \centering
            \includegraphics[width=\linewidth]{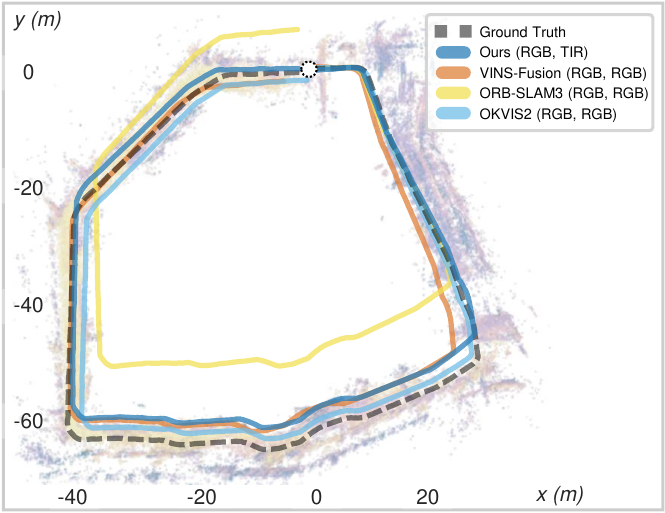}
            \caption{
                \textbf{Qualitative trajectory on INOUT-TRANS.} \bl{This sequence includes an indoor-to-outdoor transition with abrupt illumination changes between a dark indoor environment and a brighter outdoor scene. Trajectories are visualized after aligning the initial pose and starting point.}
            }
            \label{fig:trajectory_inout_trans}
        \end{minipage}
        \hfill
        \begin{minipage}[t]{0.615\linewidth}
            \centering
            \includegraphics[width=\linewidth]{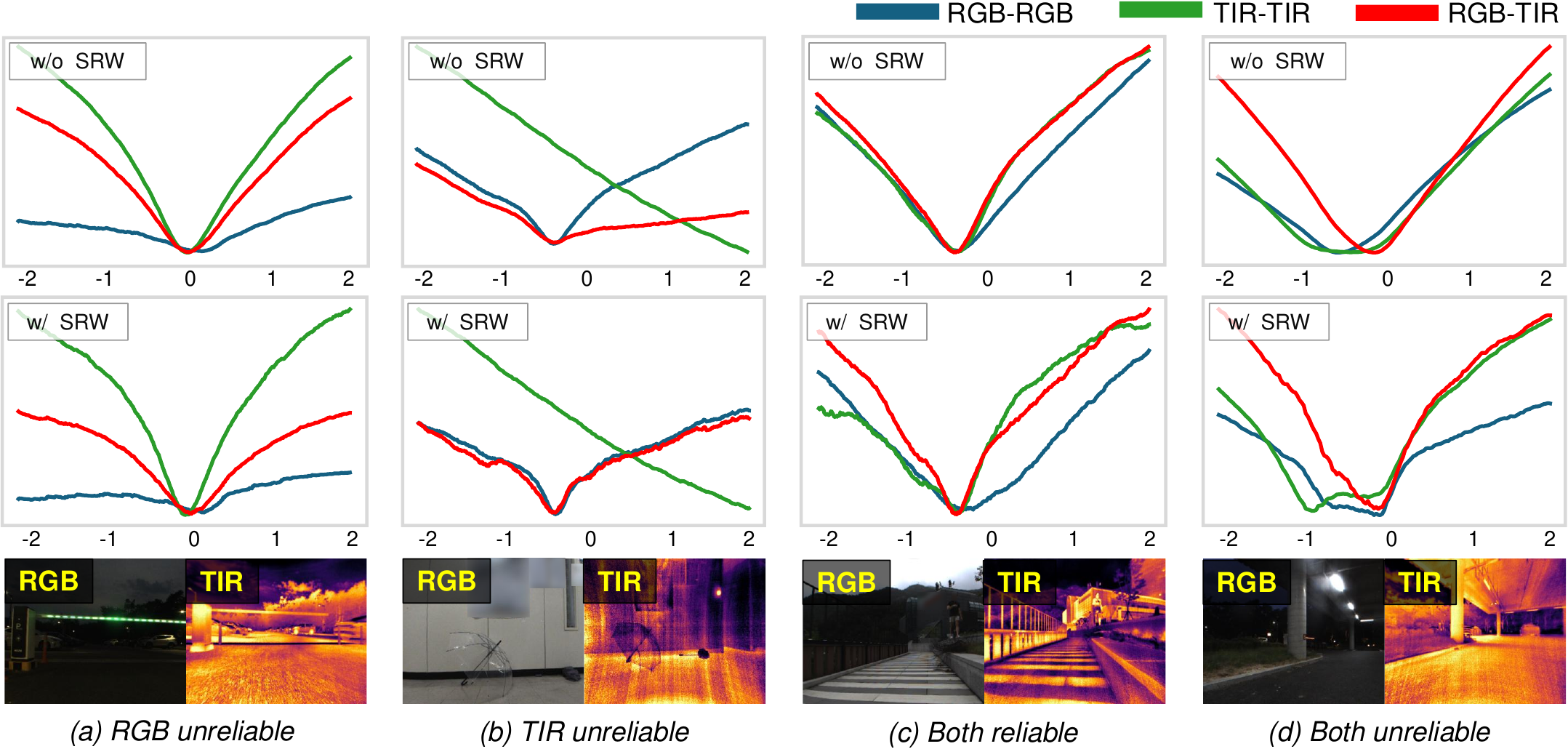}
            \caption{
                \textbf{Photometric residual landscape w.r.t.\ yaw perturbation.} \bl{The x-axis denotes yaw perturbation from ground truth, and the y-axis denotes normalized photometric error.} The plots compare RGB-only, TIR-only, and fused RGB-TIR configurations. \bl{SRW fusion (bottom) tends to suppress unreliable modalities and produces a more stable basin.}
            }
            \label{fig:residual_plot}
        \end{minipage}
    \end{minipage}
    \vspace{-6mm}
\end{figure*}

This result is particularly important because it demonstrates that the benefit of cross-spectral fusion is not limited to degraded environments. The residual landscape analysis in \figref{fig:residual_plot} provides further insight: when both visible and thermal measurements are reliable, the RGB-TIR configuration produces a sharper and more convex basin than RGB-RGB. This demonstrates that, even in nominal environments, adding thermal measurements sharpens the residual shape by providing complementary constraints rather than duplicating redundant spectral information as in a homogeneous stereo setup.

Overall, the results are consistent with our central claim that homogeneous stereo suffers from spectral redundancy, where duplicated spectral information yields correlated gradients and limited conditioning gain. RGB-TIR fusion breaks this redundancy by introducing complementary constraints that tighten the optimization landscape, improving accuracy even in nominal conditions.

\begin{table*}[h]
\vspace{1.5mm}
    \centering
    \caption{\textbf{Quantitative Comparison on INOUT-TRANS and TIR-NOISE-IN.} \ac{ATE} and \ac{RTE} are reported. Best and second-best results are shown in \textbf{bold} and \underline{underlined}, respectively. \textit{TF}/\textit{IF} denote tracking/initialization failure.}
    \label{tab:results_inout_noise}

    \setlength{\tabcolsep}{2.5pt} 
    \renewcommand{\arraystretch}{1.4}
    \setlength{\aboverulesep}{0pt}
    \setlength{\belowrulesep}{0pt}

    \resizebox{0.98\linewidth}{!}{%
        \scriptsize
        \begin{tabular}{cc | c ccc ccc c >{\columncolor{black!10}}c | c ccc ccc c >{\columncolor{black!10}}c}
        \toprule
        
        \multicolumn{2}{c|}{} 
        & \multicolumn{9}{c|}{\texttt{INOUT-TRANS}} 
        & \multicolumn{9}{c}{\texttt{TIR-NOISE-IN}} \\
        
        \cmidrule(lr){3-11} \cmidrule(l){12-20}
        
        \multicolumn{2}{c|}{}
        & \multicolumn{1}{c}{\bl{\textbf{Mono RGB}}}
        & \multicolumn{3}{c}{\textbf{Stereo RGB}} 
        & \multicolumn{3}{c}{\textbf{Stereo TIR}} 
        & \multicolumn{2}{c|}{\textbf{Cross-spectral}} 
        & \multicolumn{1}{c}{\bl{\textbf{Mono RGB}}}
        & \multicolumn{3}{c}{\textbf{Stereo RGB}} 
        & \multicolumn{3}{c}{\textbf{Stereo TIR}} 
        & \multicolumn{2}{c}{\textbf{Cross-spectral}} \\ 
        
        \cmidrule(lr){3-3} \cmidrule(lr){4-6} \cmidrule(lr){7-9} \cmidrule(lr){10-11}
        \cmidrule(lr){12-12} \cmidrule(lr){13-15} \cmidrule(lr){16-18} \cmidrule(l){19-20}
        
        \multicolumn{2}{c|}{\textbf{Metric}}
        & \bl{\tiny DM-VIO}
        & \tiny VINS-Fusion & \tiny ORB-SLAM3 & \tiny OKVIS2 
        & \tiny VINS-Fusion & \tiny ORB-SLAM3 & \tiny OKVIS2 
        & \tiny ROVTIO & \tiny \textbf{Ours} 
        & \bl{\tiny DM-VIO}
        & \tiny VINS-Fusion & \tiny ORB-SLAM3 & \tiny OKVIS2 
        & \tiny VINS-Fusion & \tiny ORB-SLAM3 & \tiny OKVIS2 
        & \tiny ROVTIO & \tiny \textbf{Ours} \\ 
        \midrule
        
        \multirow{2}{*}{\textbf{ATE}} 
          & \textbf{Rot} (deg)
          & \bl{15.294}
          & \underline{2.690} & 2.859 & 6.976 & 2.883 & 4.819 & 7.079 & 7.520 & \textbf{2.309}
          & \bl{\textit{IF}}
          & 2.993 & \textbf{2.510} & \underline{2.712} & 14.461 & \textit{IF} & \textit{TF} & \textit{TF} & 4.597 \\
         
          & \textbf{Trans} (m)
          & \bl{12.333}
          & 2.214 & 5.434 & \underline{1.710} & 2.251 & 6.908 & 85.374 & 13.282 & \textbf{1.485}
          & \bl{\textit{IF}}
          & \underline{0.405} & 0.571 & \textbf{0.397} & 6.832 & \textit{IF} & \textit{TF} & \textit{TF} & 0.896 \\
        
        \cmidrule{3-20}
        
        \multirow{2}{*}{\textbf{RTE}}
          & \textbf{Rot} (deg)
          & \bl{\textbf{0.456}}
          & 0.719 & \underline{0.499} & 1.088 & 0.721 & 0.518 & 1.027 & 0.522 & 0.471
          & \bl{\textit{IF}}
          & 1.003 & \textbf{0.265} & 1.151 & 1.174 & \textit{IF} & \textit{TF} & \textit{TF} & \underline{0.335} \\
         
          & \textbf{Trans} (m)
          & \bl{0.088}
          & \underline{0.045} & 0.050 & 0.047 & \underline{0.045} & 0.048 & 0.432 & 0.123 & \textbf{0.033}
          & \bl{\textit{IF}}
          & 0.037 & \textbf{0.021} & 0.035 & 0.172 & \textit{IF} & \textit{TF} & \textit{TF} & \underline{0.022} \\
        
        \bottomrule
        \end{tabular}%
    }
    \vspace{-4mm}
\end{table*}

\paragraph{\texttt{NIGHT-OUT}}
This sequence is characterized by prolonged low-light and near-darkness conditions. As reported in \tabref{tab:results_night_glare}, our method achieves the lowest translational \ac{ATE} and \ac{RTE}. Interestingly, due to intermittent street lamps, the RGB-stereo configuration remains relatively robust in this sequence, sometimes outperforming TIR-stereo in terms of stability. The localized artificial lighting provides sufficient gradients for short-term tracking, partially compensating for the overall darkness. In contrast, the TIR-stereo configuration suffers from amplified noise due to the relatively uniform evening temperature and reduced thermal contrast in indoor segments. Consequently, \ac{FPN} and noise-dominated gradients degrade depth consistency and scale stability.

In this challenging setting, the proposed \ac{SRW} plays a role. By adaptively down-weighting noise-dominated thermal residuals while preserving informative cross-spectral constraints, \ac{SRW} prevents corrupted \ac{TIR} measurements from destabilizing the optimization. Consequently, our cross-spectral system achieves the lowest translational errors, demonstrating stable scale recovery even when one modality becomes unreliable.

\paragraph{\texttt{GLARE-OUT}}
This sequence is dominated by persistent over-exposure that affects large portions of the image. As reported in \tabref{tab:results_night_glare}, RGB baselines suffer from pronounced scale drift under glare, resulting in large translation errors, while \ac{TIR} pipelines remain sensitive to noise and texture sparsity. Contrastingly, our method achieves the lowest translation and rotational error among all compared methods, indicating stable local motion estimation under sustained over-exposure, as thermal structures provide reliable geometric cues and cross-spectral depth constraints mitigate correlated RGB failures. ROVTIO~\cite{flemmen2021rovtio}, which does not employ direct cross-modal matching, exhibits substantially larger scale drift, highlighting the importance of explicitly coupling RGB and thermal observations through cross-spectral constraints.

\paragraph{\texttt{INOUT-TRANS}}
This sequence captures a transition from a dark indoor parking structure to an outdoor environment. Most RGB baselines experience early scale degradation while traversing the dim indoor segment, which subsequently propagates into large translational errors after exiting to the outdoor scene, as shown in \tabref{tab:results_inout_noise}. In contrast, \ac{TIR} pipelines exhibit severe scale drift throughout the sequence, reflecting their limited ability to adapt to changing scene geometry and thermal characteristics across indoor and outdoor environments. Our method remains robust throughout the transition and achieves the lowest errors across all evaluation metrics. These results highlight that cross-spectral RGB-TIR fusion offers stronger adaptability to evolving environmental conditions by mitigating scale drift in single-modality pipelines and maintaining consistent geometric constraints across heterogeneous scenes.

\begin{table*}[t]
\vspace{1.5mm}
    \centering
    \def\TotalWidth{0.96} 

    \begin{minipage}{\TotalWidth\linewidth}
        \centering
        \begin{minipage}[c]{0.42\linewidth}
            \centering
            \caption{\textbf{Ablation of Spectral Reliability Weight (SRW).} Comparison of naive fusion (w/o SRW) and the full system (w/ SRW), showing that SRW suppresses modality-specific artifacts, reduces \ac{ATE}, and improves robustness. Best results are in \textbf{bold}.}
            \label{tab:ablation_srw}
    
            \setlength{\tabcolsep}{3.5pt} 
            \renewcommand{\arraystretch}{1.3}
            
            \resizebox{\linewidth}{!}{%
                \begin{tabular}{cc | c >{\columncolor{black!10}}c | c >{\columncolor{black!10}}c}
                \toprule
                \multicolumn{2}{c|}{} 
                & \multicolumn{2}{c|}{\texttt{DAY-OUT}} 
                & \multicolumn{2}{c}{\texttt{GLARE-OUT}} \\
                
                \cmidrule(lr){3-4} \cmidrule(l){5-6}
                
                \multicolumn{2}{c|}{\textbf{Metric}} 
                & w/o SRW & \textbf{w/ SRW} & w/o SRW & \textbf{w/ SRW} \\
                \midrule
                
                \multirow{2}{*}{\textbf{ATE}} 
                  & \textbf{Rot} (deg)
                  & 2.821 & \textbf{2.269} & 2.991 & \textbf{2.058} \\
                  & \textbf{Trans} (m)
                  & 0.350 & \textbf{0.234} & 0.544 & \textbf{0.419} \\
                \cmidrule{1-6}
                \multirow{2}{*}{\textbf{RTE}}
                  & \textbf{Rot} (deg)
                  & \textbf{0.496} & 0.506 & \textbf{0.809} & 0.810 \\
                  & \textbf{Trans} (m)
                  & 0.019 & \textbf{0.018} & 0.029 & \textbf{0.026} \\
                
                \bottomrule
                \end{tabular}%
            }
        \end{minipage}
        \hfill
        \begin{minipage}[c]{0.56\linewidth}
            \centering
            \includegraphics[width=\linewidth]{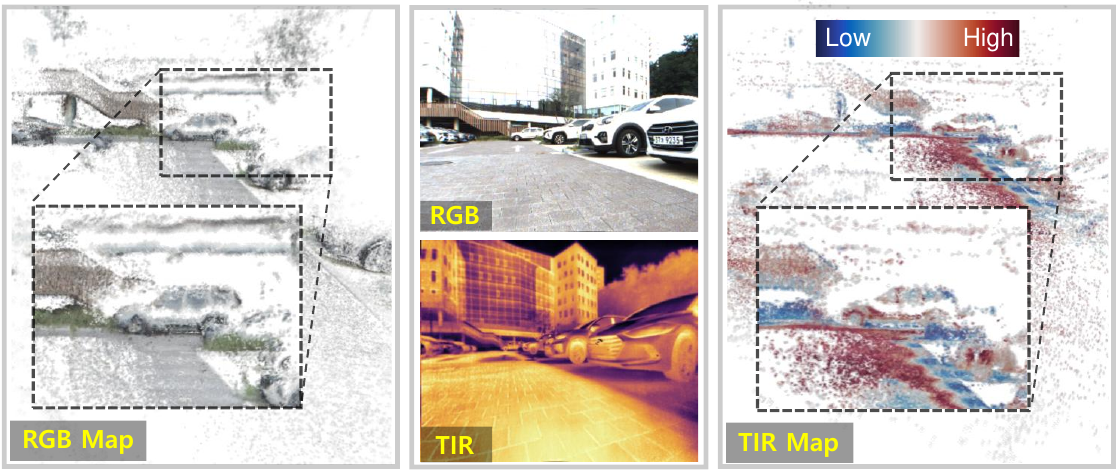}
            \captionof{figure}{\textbf{Temperature-aware 3D reconstruction.} Outdoor RGB and thermal maps, where the thermal cloud (\textcolor{blue}{Low} to \textcolor{red}{High}) reveals details invisible in RGB.
            }
            \label{fig:map}
        \end{minipage}
    \end{minipage}
    \vspace{-6mm}
\end{table*}

\paragraph{\texttt{TIR-NOISE-IN}}
This sequence is characterized by severe thermal \ac{FPN} and low-texture indoor scenes. Under such conditions, the thermal intensity distribution becomes highly compressed, amplifying sensor noise and severely degrading geometric consistency. As reported in \tabref{tab:results_inout_noise}, most TIR-based baselines either fail to initialize or encounter tracking failure. In contrast, our method successfully completes the entire sequence. Among TIR-based configurations, it demonstrates substantially improved robustness, which can be attributed to the proposed \ac{SRW}. By down-weighting noise-dominated thermal residuals, \ac{SRW} prevents \ac{FPN}-induced artifacts from corrupting the optimization and stabilizes scale estimation even in severely degraded thermal conditions. RGB-based baselines achieve lower translation errors in this sequence, as the indoor scene provides stronger visible structure than thermal contrast. Nevertheless, the proposed cross-spectral framework maintains stable tracking without initialization or tracking failure, demonstrating the effectiveness of \ac{SRW} in suppressing extreme thermal noise.

\begin{table}[t]
    \centering
    \caption{
        \textbf{Runtime statistics on \texttt{DAY-OUT}.}
        Latency values are reported in milliseconds. The camera input runs at 25 Hz(40 \texttt{ms}).
    }
    \label{tab:runtime_stats}
    \vspace{-2mm}
    \setlength{\tabcolsep}{4pt}
    \renewcommand{\arraystretch}{0.85}
    \resizebox{0.85\columnwidth}{!}{
    \begin{tabular}{l|ccc}
        \toprule
        \textbf{Module} & \textbf{Mean}\texttt{[ms]}& \textbf{Median} \texttt{[ms]}& \textbf{P95} \texttt{[ms]}\\
        \midrule
        Preprocess & 11.29 & 10.26 & 16.63 \\
        Frontend Tracking & 7.19 & 6.69 & 10.43 \\
        Sliding-window Opt. & 72.87 & 75.24 & 98.04 \\
        Async RGB-\ac{TIR} Match. & 107.64 & 106.48 & 118.06 \\
        Pose Output Period & 40.00 & 40.00 & 40.56 \\
        \bottomrule
    \end{tabular}
    }
    \vspace{-5mm}
\end{table}

\vspace{-4mm}

\subsection{Qualitative Analysis}
\label{sec:exp_qual}
We visualize estimated trajectories overlaid on the global map to assess tracking consistency. In \texttt{DAY-OUT} (\figref{fig:trajectory_day_out}), our fused estimate yields a smoother trajectory than RGB-only baselines, indicating benefits beyond degraded illumination. In \texttt{INOUT-TRANS} (\figref{fig:trajectory_inout_trans}), RGB baselines drift under abrupt lighting changes, while \ac{TIR}-only methods degrade due to indoor thermal noise, whereas our system remains well aligned with the ground truth. The reconstructed thermal point cloud (\figref{fig:map}) also reveals temperature-dependent structures, such as heated vehicle engines, that are not visible in RGB.

\vspace{-4mm}
\subsection{Cross-Spectral Fusion via Residual Landscapes}
\label{sec:exp_srw_analysis}

We analyze the robustness mechanism of the proposed system through the photometric residual landscape. Using LiDAR-derived ground-truth relative poses, we evaluate the aggregated photometric error of each modality pairing while perturbing the yaw around the ground-truth pose. Combined with the ablation results in \tabref{tab:ablation_srw}, this analysis highlights two key effects of cross-spectral fusion. First, \ac{SRW} enables selective modality reliance. As shown in Fig.~\ref{fig:residual_plot} (a), (b), when one modality is degraded, naive fusion produces a distorted or flattened basin. In contrast, \ac{SRW} suppresses noise-dominated residuals and preserves the reliable spectrum, restoring a stable optimization landscape. The ablation in \tabref{tab:ablation_srw} shows that removing \ac{SRW} can increases \ac{ATE}, confirming its role in suppressing modality-specific outliers. Moreover, heterogeneous fusion sharpens the objective: in nominal conditions (c), the RGB-TIR residual yields a tighter basin than either modality alone, and even when both are weak (d), their combination still forms a distinct minimum, reflecting complementary spectral information.
\vspace{-4mm}

\section{Conclusion}
\label{sec:conclusion}


In this paper, we presented a real-time cross-spectral \ac{VTI} odometry system that mitigates \emph{spectral redundancy} in homogeneous stereo, which leads to correlated failures under adverse illumination and limited gains in residual curvature. Our experiments demonstrate that RGB-\ac{TIR} fusion provides complementary constraints and improves robustness even under nominal daytime conditions. Real-time operation is achieved by decoupling deep RGB-\ac{TIR} matching from high-rate estimation and applying modality-aware reliability weighting during fusion. A remaining limitation is the loss of either modality: when one modality becomes unavailable, cross-spectral matching cannot be performed, and the system falls back to single-modality inertial odometry. Future work will explore RGB-\ac{TIR} place recognition and full cross-spectral SLAM.

\vspace{-4mm}

\footnotesize
\bibliographystyle{IEEEtranN}
\bibliography{string-short,root}


\end{document}